\documentclass{Interspeech2024}

\newenvironment{tight}{%
\setlength{\abovedisplayskip}{3pt}
\setlength{\belowdisplayskip}{3pt}
}

\usepackage{CJKutf8}
\usepackage{CJK}
\usepackage{tabularx}
\usepackage[utf8]{inputenc}

\interspeechcameraready

\title{Using Large Language Model for End-to-End Chinese ASR and NER}

\name{Yuang Li$^{1*}$, Jiawei Yu$^{2*}$, Min Zhang$^1$, Mengxin Ren$^1$, Yanqing Zhao$^1$, \\ Xiaofeng Zhao$^{1}$, Shimin Tao$^1$, Jinsong Su$^2$, Hao Yang$^1$}

\address{
 $^{1}$Huawei Translation Services Center, Beijing, China\\
 $^{2}$School of Informatics, Xiamen University, China
}

\email{
liyuang3@huawei.com, yujiawei@stu.xmu.edu.cn, jssu@xmu.edu.cn, yanghao30@huawei.com}

\keywords{speech recognition, named entity recognition, large language model}

\newcommand\blfootnote[1]{%
  \begingroup
  \renewcommand\thefootnote{}\footnote{#1}%
  \addtocounter{footnote}{-1}%
  \endgroup
}

\begin{document}

\maketitle

\begin{abstract}



Mapping speech tokens to the same feature space as text tokens has become the paradigm for integrating speech modality into decoder-only large language models (LLMs). An alternative is to use an encoder-decoder architecture that incorporates speech features through cross-attention. In this work, we connect the Whisper encoder with ChatGLM3 and provide in-depth comparisons of these two approaches using Chinese automatic speech recognition (ASR) and named entity recognition (NER) tasks. We evaluate their performance using the F1 score and a fine-grained taxonomy of ASR-NER errors. Our experiments reveal that the encoder-decoder model outperforms the decoder-only model if the context is short, while the decoder-only model benefits from a long context as it fully exploits all layers of the LLM. Additionally, we obtain a state-of-the-art F1 score of 0.805 on the AISHELL-NER test set by using chain-of-thought NER which first infers long-form ASR transcriptions and then predicts NER labels.

\end{abstract}

\section{Introduction}

Large language models (LLMs) have been shown to perform remarkably on natural language processing tasks, such as question answering, summarization, and machine translation~\cite{gpt2023gpt}. Various approaches have been proposed to leverage the power of LLMs to multi-modalities. Early works focus on visual understanding tasks. MiniGPT-4~\cite{zhu2023minigpt} directly feeds visual features into the LLM through a projection layer. LLaMA-Adapter~\cite{zhang2023LLaMA} adopts fixed-length trainable vectors as layer-wise prompts which can include visual information. MiniGPT-4 and LLaMA-Adapter are decoder-only models, whereas Flamingo~\cite{alayrac2022flamingo} utilizes an encoder-decoder framework where visual representations are merged into the LLM through cross-attention. 
\blfootnote{$^\star$ Equal contribution.}

Recently, combining speech encoders with LLMs has gained momentum. Among various applications, the automatic speech recognition (ASR) task has received the most attention~\cite{li2023prompting, ling2023adapting, yu2023connecting, wu2023decoderonly, fathullah2023prompting, hono2023integration}. Most existing works concentrate on the Adapter layer, which is used for reducing the dimensionality of the speech features and mapping them to the text embedding space. Different types of Adapter layers have been proposed, such as the Attention layer~\cite{yu2023connecting}, the adaptive CTC downampler~\cite{ling2023adapting}, and the Convolutional layers~\cite{fathullah2023prompting}. Moreover, a variety of speech encoders (e.g., Whisper encoder~\cite{radford2022robust} and HuBERT~\cite{hsu2021hubert}) and LLMs (e.g., LLaMA~\cite{touvron2023llama} and Vicuna~\cite{vicuna2023}) have been explored in this context. Beyond ASR, the potential of LLMs has been further unleashed by applying them to more challenging tasks, such as speech translation~\cite{chen2023salm, huang2023speech}, ASR error correction~\cite{Radhakrishnan_2023}, and multitask speech and audio event understanding~\cite{gong2023joint, liang2023acoustic, wang2023slm, chu2023qwenaudio}. However, these methods adopt a decoder-only architecture that takes speech or audio features as the input to LLMs (similar to miniGPT4), which differs from the standard encoder-decoder architecture of ASR~\cite{chorowski2015attention}. The only exception is~\cite{li2023prompting}, where the HuBERT speech encoder is integrated with the LLaMA model via cross-attention for ASR domain adaptation. In this study, we conducted a thorough comparison of the two types of architectures on Chinese ASR and named entity recognition (NER) tasks, which have received less attention in prior research.

NER from speech is a fundamental task in spoken language understanding, which aims to identify and classify named entities into predefined categories, such as person (PER), location (LOC), and organization (ORG). This task can be performed by either a pipeline system~\cite{interspeech/JannetGAR17/pipeline-metric-tune} or an end-to-end (E2E) system~\cite{DBLP:journals/corr/abs-2210-11987,DBLP:conf/icassp/SerdyukWFKLB18,DBLP:conf/interspeech/YadavG0S20/english, DBLP:conf/interspeech/MdhaffarDPE22/france,DBLP:conf/emnlp/AroraDYMB022-findings}. A pipeline system consists of an ASR module and a text-based NER model, where the input audio is first transcribed by the ASR module, and then the resulting ASR output is processed by the NER model. In contrast, an E2E system directly extracts entities from speech without depending on the intermediate ASR output, therefore avoiding error propagation. We chose the NER task for our experiments because it requires the LLM to not only learn the mapping from speech features to text tokens, but also comprehend the semantic meaning of the ASR transcription. To analyze the ASR-NER results in detail, we applied the taxonomy of ASR-NER errors proposed in \cite{DBLP:conf/acl/SzymanskiAMSSZ23} for the pipeline system to our LLM systems.


In this paper, we combined the Whisper encoder~\cite{radford2022robust} with the ChatGLM3-6B~\cite{zeng2023glm130b}. The Whisper is an ASR model trained on a large-scale speech corpus of 680k hours of data and ChatGLM-6B is a bilingual LLM that has outstanding performance in Chinese. We only fine-tuned the LoRA~\cite{hu2022lora} adapters and the connectors between the Whisper encoder and ChatGLM3-6B on AISHELL datasets~\cite{aishell, Chen_2022}. Through extensive experiments and visualizations of gate values and attention scores, we uncovered that the encoder-decoder architecture with cross-attention leverages the deeper layers of the LLM and achieves superior performance on the short-form ASR task, while the decoder-only architecture with self-attention exploits all the LLM layers and excels on the long-context ASR and NER tasks. Additionally, we employed chain-of-thought (CoT) NER, which first generates long-form ASR transcriptions and then predicts NER labels. CoT NER attained a state-of-the-art (SOTA) F1 score of 0.805 on the AISHELL-NER test set~\cite{Chen_2022}. According to the taxonomy of ASR-NER results, CoT NER achieved an absolute reduction in omission errors by 7\% and an absolute improvement of 9\% in correct entities compared to the baseline Conformer model.

\begin{figure}[h]
  \centering
  \includegraphics[width=\linewidth]{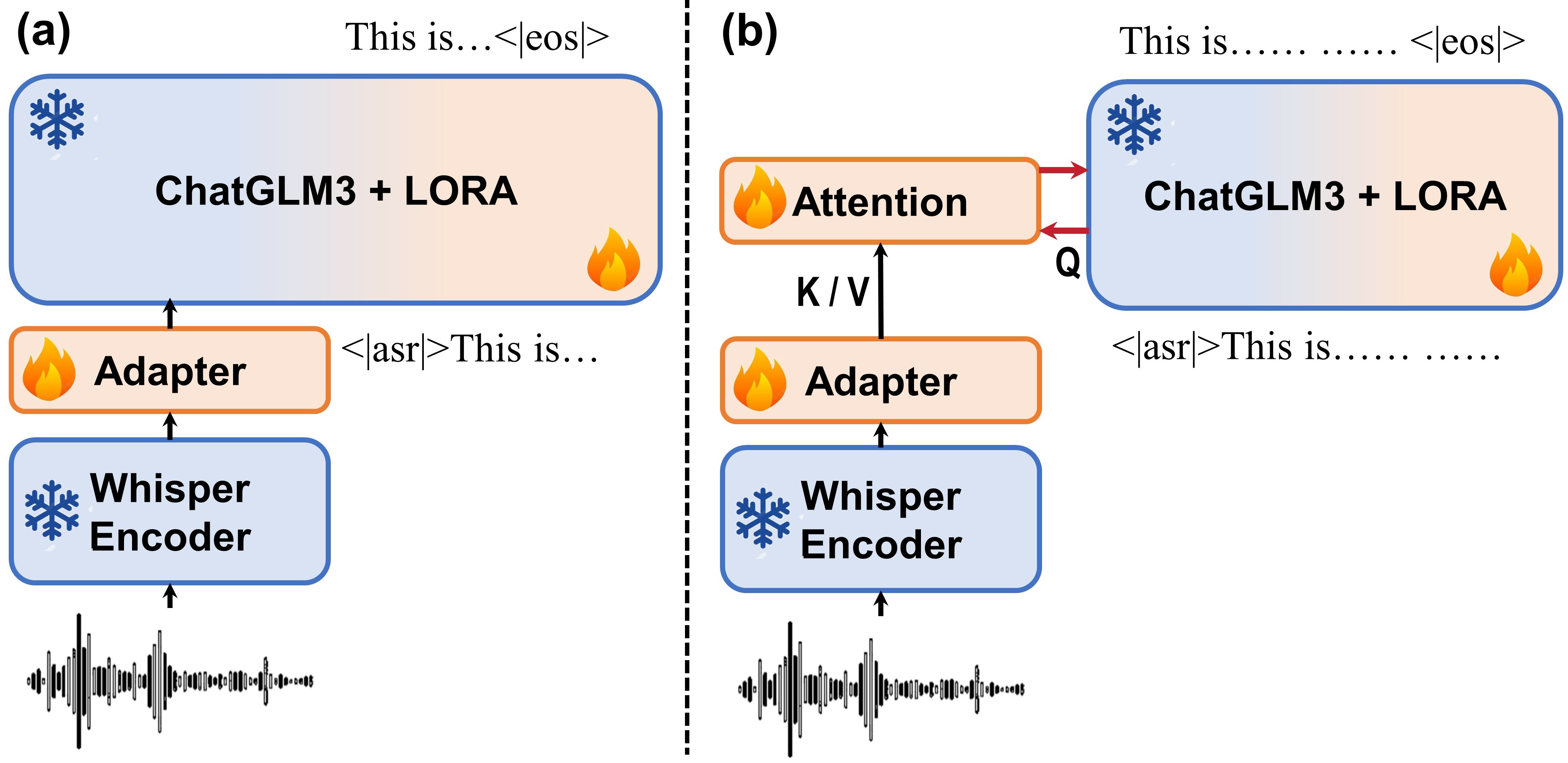}
  \caption{The speech modality is incorporated into the LLM through (a) an adapter (decoder-only), and (b) cross-attention layers (encoder-decoder).}
  \label{fig:arch}
\end{figure}

\section{Methodology}

\subsection{Decoder-only model}


One simple way to incorporate the speech modality into the LLM is to use an Adapter layer that bridges the gap between the speech features and the text embeddings. Figure~\ref{fig:arch} (a) illustrates how the ChatGLM3 model takes the speech tokens from the Whisper encoders, which are transformed by an Adapter, as the input and generates ASR transcriptions in an autoregressive way. In this setting, the speech tokens act as prompts. Since speech features usually have much longer lengths than text features, we downsample the speech features by stacking every five adjacent frames. Afterwards, the speech features are fed into two linear layers as shown in Equation~\ref{eq:1}.

 \begin{tight}
 \begin{align}
 \label{eq:1}
 \mathbf{S} = \text{Linear}(\text{ReLU}(\text{Linear}(\mathbf{H}_{whisper})))
\end{align}
\end{tight}

\noindent where $\textbf{H}_{whisper}$ is the downsampled speech features and $\textbf{S}$ is the speech tokens, the output of the Adapter layer.

\subsection{Encoder-decoder model}


Figure~\ref{fig:arch} (b) shows the integration of the Whisper encoder into ChatGLM3, which follows the traditional Transformer~\cite{NIPS2017_3f5ee243} architecture where the encoder and the decoder are connected through cross-attention. After each self-attention layer of ChatGLM3, a cross-attention layer is added where the hidden states of text tokens serve as queries and the speech features serve as keys and values. Similar to the approach in~\cite{alayrac2022flamingo, li2023prompting}, we adopt gated cross-attention with learnable gates to control the amount of influence that the speech modality has on the final output. The gate values are initialized to zeros to stabilize training. Unlike previous works, we swap the order of the feedforward and cross-attention layers and apply the gates only to the cross-attention layers (Equation~\ref{eq:2},~\ref{eq:3}, and~\ref{eq:4}). In our initial experiments, this improves the training stability and the performance of our model, as the speech features are processed by more layers before being fed into the LLM. 

 \begin{tight}
 \begin{align}
 \label{eq:2}
\mathbf{S} &= \text{ReLU}(\text{Linear}(\mathbf{H}_{whisper}))\\
\label{eq:3}
\mathbf{S}^{(i)} &= \text{ReLU}(\text{Linear}^{(i)}(\mathbf{S}))\\
\label{eq:4}
\mathbf{H}^{(i)} = \mathbf{H}^{(i)} + & \text{Tanh}(g^{(i)}) \odot \text{XATT}^{(i)}(\mathbf{H}^{(i)}, \mathbf{S}^{(i)}, \mathbf{S}^{(i)})
\end{align}
\end{tight}

\noindent where the downsampled speech features $\textbf{H}_{whisper}$ are fed into a single linear layer, resulting in new features $\textbf{S}$. Then at the $i_{th}$ layer, $\textbf{S}$ is processed by a Linear layer followed by a cross-attention (XATT) layer scaled by a Tanh gate.

The main differences between the Adapter and the cross-attention architectures can be summarized as follows:

\begin{itemize}
\item \textbf{Principle}: The Adapter layer enables the LLM to handle speech tokens and text tokens uniformly. The cross-attention layer considers the speech features as a distinct source sequence from the text tokens.

\item \textbf{Implementation}: The Adapter architecture uses the speech tokens as inputs whereas the cross-attention architecture incorporates the speech features after each layer. Consequently, the Adapter architecture has the advantage of requiring less modification to the source code of the LLM.

\item \textbf{Parameters and Computation}: Compared to the naive Adapter method, the cross-attention approach introduces a much larger number of parameters, since it involves cross-attention at each layer. Nevertheless, we found that the cross-attention architecture achieves faster training and inference because the cross-attention operates at a lower dimension than the self-attention in the LLM.
\end{itemize}

\subsection{Long-form ASR and CoT NER}

We propose a three-phase training schedule to adapt our Whisper-ChatGLM3 models to Chinese ASR and NER. The training involves three tasks including short-form ASR, long-form ASR, and CoT NER. Table~\ref{tab:task} illustrates the input formats of these tasks, and the details are provided as follows:

\begin{itemize}
     
\item \textbf{Short-form ASR}: We select a variable number of utterances at random and concatenate their features and their corresponding transcriptions. A special token, denoted by $|\textbf{asr}|$, is used to indicate the ASR task. Random concatenation was shown to be effective for the Whisper model~\cite{yu2023connecting}, as it uses 30-second inputs that are longer than the typical segments in the ASR corpus.

\item \textbf{Long-form ASR}: The use of historical context from both speech and text modalities was shown to enhance ASR performance~\cite{schwarz2020improving, hori2021advanced}. Motivated by this, we investigated the effect of incorporating both historical speech and text information in the recognition of the current utterance. We construct an input sequence by concatenating historical speech tokens $S_{i-1}$ and current speech tokens $S_{i}$, separated by a special token $\gamma$ that marks the beginning of the current utterance. The model produces a long-form transcription with a special token $|\textbf{sep}|$ that indicates the start of the current transcription.

\item \textbf{CoT NER}: We instruct the LLM to produce long-form transcriptions first and then assign NER labels to the current utterance. This approach can be regarded as a CoT process. It can also be viewed as the combination of pipeline and E2E NER systems as the model accesses both ASR transcriptions and speech features.

\end{itemize}

\begin{table}[h]
\small
\caption{Different training tasks. $S$ and $T$ denote speech and text tokens for an utterance respectively. $|\cdot|$ indicates special text tokens. $\gamma$ is a special audio token that separates the current and historical speech tokens.}
    \centering
    \begin{tabular}{p{2.1cm} | p{5cm}} 
    \toprule
    Short-form ASR & $S_i, S_j|\textbf{asr}|T_i, T_j$\\
    Long-form ASR & $S_{i-1}, \gamma, S_i|\textbf{asr}|T_{i-1}|\textbf{sep}|T_{i}$\\
    CoT NER & $S_{i-1}, \gamma, S_i|\textbf{asr}|T_{i-1}|\textbf{sep}|T_{i}|\textbf{ner}|\hat{T}_{i}$\\
    \bottomrule
    \end{tabular}
    \label{tab:task}
\end{table}

In the first two phases of training, the models are optimized on short-form and long-form ASR tasks respectively. In the third phase, we perform multitask training of long-form ASR and CoT NER jointly.

\subsection{Categories of NER predictions}
Referring to the method in~\cite{DBLP:conf/acl/SzymanskiAMSSZ23}, we conducted a fine-grained analysis of NER predictions. This allows us to gain a deeper insight into the sources of errors, the benefits of LLMs, and the impact of historical information on NER performance. The categories of NER predictions include:

\begin{itemize} 
\item \textbf{Correct Span}: Entities that match the gold entity tags, meaning the location of the entity is accurately predicted in the ASR transcription.
\item \textbf{Correct Entity}: A subset of correct span. All the tokens within the entity are predicted accurately. This is the only category that is considered correct by the standard F1 score. 
\item \textbf{Error Span}: Entities that deviate from the gold entity tags. 
\item \textbf{Replacement}: A subset of error span. The entity type is predicted incorrectly. 
\item \textbf{Omission}: A subset of error span. The entity in the gold transcript is missing. 
\end{itemize}

\section{Experimental Setups}

\subsection{Dataset and metrics}
In our experiments, we used the AISHELL-NER dataset~\cite{Chen_2022}, an annotated version of the AISHELL-1~\cite{aishell} dataset, which consists of 170 hours of Chinese speech data. The training set of AISHELL-NER contains about 120,000 sentences, and the test set has 7176 sentences. The dataset annotates three types of named entities: person (PER), location (LOC), and organization (ORG), using special symbols: “$[\cdot]$” for PER, “$(\cdot)$” for LOC, and “$<\cdot>$” for ORG. We formulated NER as a sequence generation task, where the model predicts the entity symbols along with the ASR transcriptions. We measured the performance of our model on the test set, which has 900 PER, 1330 LOC, and 1165 ORG entities, using character error rate (CER) for ASR and F1 score for NER.

\subsection{Model architecture}
We employed the Conformer-based~\cite{Chen_2022} E2E system as a baseline model that did not incorporate the LLM. Our systems utilized the encoder of Whisper-large-v2~\cite{radford2022robust} as the speech encoder and ChatGLM3-6B~\cite{zeng2023glm130b} as the base LLM. We froze the Whisper encoder and applied efficient fine-tuning of ChatGLM3 with LoRA~\cite{hu2022lora}, enabling the LLM to adapt to the domain of the AISHELL dataset and comprehend ASR and NER tasks. The fine-tuning of the LLM avoided reliance on the parameter in the Adapter to accomplish the task that the language model should handle. For LoRA, we set the rank to 32 and applied it to both the attention and feedforward layers, resulting in 15 million parameters. For the decoder-only architecture with the Adapter layer, the Adapter had 43 million parameters with two linear layers that mapped the feature from the dimension of 6400 ($1280 \times 5$) to 4096. For the encoder-decoder architecture with the gated cross-attention, the speech features were first projected to the dimension of 4096 and then reduced to 1024 at each layer. The cross-attention layer had eight heads, and the multi-head dot-product attention was computed at a low dimension of 1024 and then projected back to 4096 after the attention layer. The linear layers and the cross-attention layers had 437 million parameters in total.

\subsection{Training schedule}
The model underwent three phases of training. In the first phase, the model was trained on the short-form ASR task for 40 epochs, using a learning rate of 5e-5 and a batch size of 64. In the second phase, the model was fine-tuned on the long-form ASR task for 20 epochs, using a learning rate of 3e-5. The number of historical utterances was randomly sampled for each training example. The concatenated audio was either padded or cropped to 30 seconds, depending on whether it was shorter or longer than that duration. In the third phase, the model was further trained for 20 epochs, using a learning rate of 3e-5. To activate the model's NER capability without compromising its ASR capability, we optimized the model for the long-form ASR task with a probability of 30\% and for the NER task with a probability of 70\%. 


\section{Experimental Results}

\subsection{ASR results}

As demonstrated in Table 2, our systems achieved a substantial improvement over the Conformer baseline, with a relative CER reduction of 19.7\%. The encoder-decoder architecture outperformed the decoder-only architecture when no historical context was available. Incorporating contextual information led to consistent ASR accuracy enhancement for both architectures. Nevertheless, the decoder-only architecture benefited more from historical information than the encoder-decoder architecture, as evidenced by the relative CER reduction of 14.4\% versus 7.0\%, respectively. This suggests that the decoder-only architecture can leverage historical information more effectively.

\begin{table}[h]
\caption{The ASR performance of two systems with or without history context measured by \textbf{CER (\%)}.}
\label{tab:asr}
\centering
\begin{tabular}{c | c c} 
\toprule
 & Short-form & Long-form\\
\midrule
Conformer & 4.83 & /\\
\midrule
Decoder-only & 4.02 & \textbf{3.44}\\
Encoder-decoder & \textbf{3.88} & 3.61\\
\bottomrule
\end{tabular}
\end{table}

\subsection{NER results}

\begin{table}[b]
\caption{The NER performance of two systems with different \textbf{historical context lengths (His.)} measured by \textbf{F1 score}.}
\label{tab:ner1}
\centering
\begin{tabular}{c | c | c c c | c} 
\toprule
 Method & His. & PER & LOC & ORG & Total\\
\midrule
Conformer & 0 & 0.561 & 0.832 & 0.778 & 0.743 \\
\midrule
Decoder-& 0 & 0.601 & 0.868 & 0.812 & 0.778 \\
only & 1 & 0.631 & 0.881 & 0.837 & 0.799 \\
& 2 & 0.635 & 0.878 & 0.853 & \textbf{0.805} \\
\midrule
Encoder- & 0 & 0.596 & 0.879 & 0.813 & 0.782 \\
decoder & 1 & 0.594 & 0.885 & 0.832 & \textbf{0.789} \\
& 2 & 0.600 & 0.878 & 0.831 & 0.788 \\
\bottomrule
\end{tabular}
\end{table}

As shown in Table~\ref{tab:ner1}, the decoder-only and encoder-decoder models achieved the highest F1 scores of 0.805 and 0.789 respectively when using historical context, which were significantly higher than the Conformer model's score of 0.743. Regarding the types of named entities, all models performed poorly on names. This is mainly because Chinese names have many variations and homophones that can create confusion and ambiguity for the ASR system. The trend of NER performance when using historical context was similar to that of ASR, in that the decoder-only model's F1 score increased steadily, while the encoder-decoder model's F1 score reached its peak with only one historical utterance.

\subsection{Visualizations}

\begin{figure}[]
  \centering
  \includegraphics[width=\linewidth]{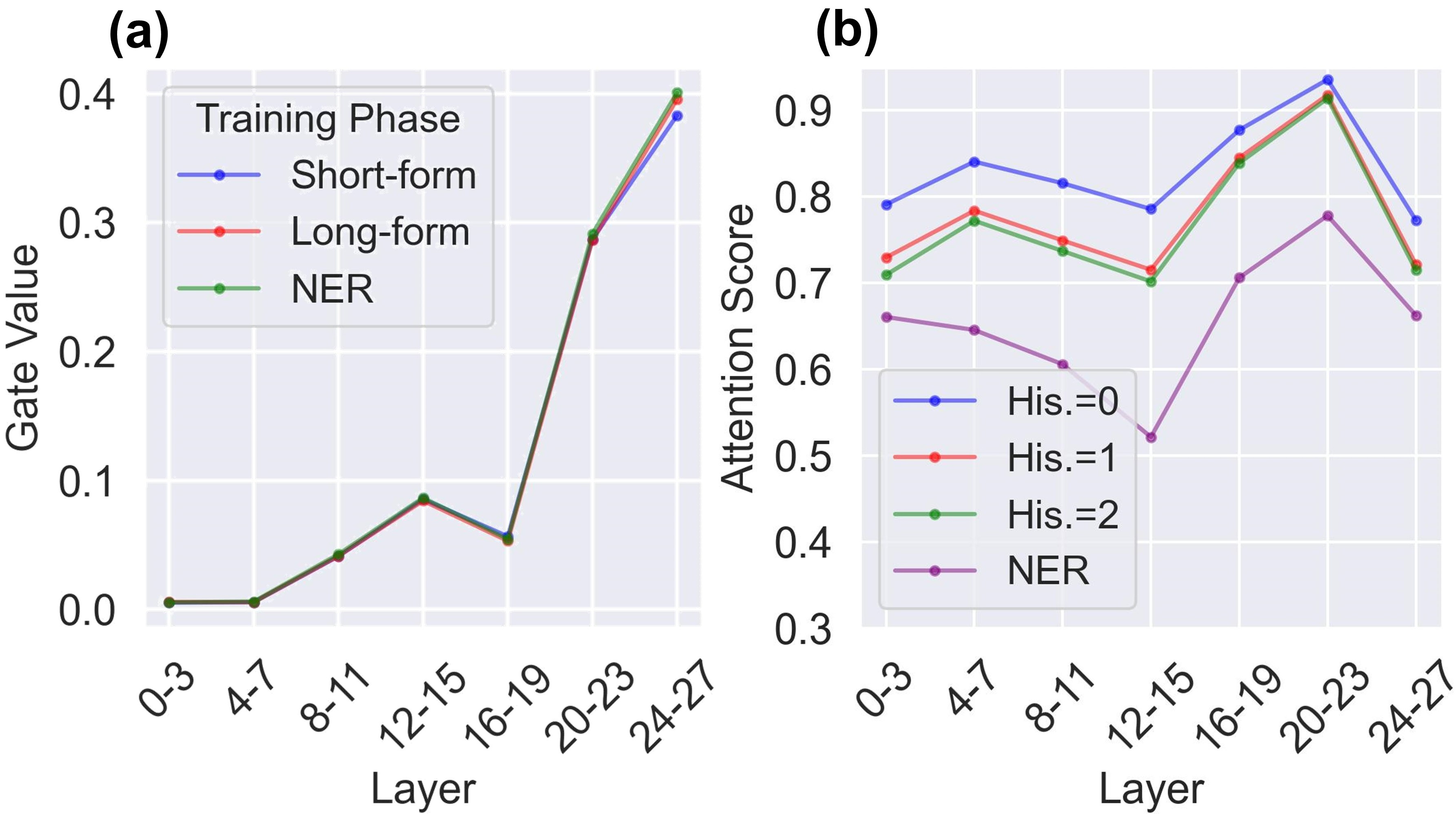}
  \caption{(a) Gate values of the cross-attention across different layers for the encoder-decoder architecture during different training phases. (b) The attention scores correspond to the speech tokens across different layers for the decoder-only architecture with different historical (His.) context lengths and tasks (i.e., ASR or NER).}
  \label{fig:vis}
\end{figure}

Figure~\ref{fig:vis} provides visualizations to better understand the results in the previous sections and the differences between the two architectures. Figure~\ref{fig:vis} (a) illustrates the gate values across different layers for the cross-attention layer. The gate values remained roughly unchanged across different training phases and deep layers were assigned significantly larger gate values than shallow layers. 

For the decoder-only architecture, the influence of speech modality is controlled by self-attention. We calculated the average attention scores for the speech tokens on the AISHELL test set (Table~\ref{fig:vis} (b)). The attention score on the speech tokens decreased with longer context as more attention was given to the text tokens. For the NER task, we first generated ASR transcriptions that provided rich semantic information for NER, resulting in the lowest attention scores for speech tokens. The attention scores are consistently above 0.5 across different layers, indicating that the speech features are important for ASR and NER. Moreover, the attention scores for the deep and shallow layers are similar, implying that all LLM layers were fully utilized. 

Based on the previous observations, we can gain insights into why the decoder-only model performed better with a longer context. First, the decoder-only model better utilized the pre-trained parameters inside the LLM by incorporating speech features at shallow layers. Additionally, the decoder-only model can dynamically adjust the importance of text tokens according to the context length and the nature of the task, while the encoder-decoder architecture treated the speech modality in a static manner with similar weights under different scenarios.

\vspace{-0.3cm}

\subsection{NER taxonomy}

Table~\ref{tab:ner2} presents the taxonomy of NER predictions. It is evident that all models have a low rate of replacement errors, whereas omission errors are more prevalent. These errors mainly stemmed from PER and ORG, which are often rare entities. One of the main advantages of using the LLM is that it reduced the omission error by almost 50\% compared to the Conformer baseline. With the LLM, more than 90\% of entities can be labeled accurately, but among them, 10\% have erroneous ASR results which are mostly substitution errors for personal names. These results prove that with the aid of the LLM, the model can better identify rare named entities, but it remains difficult to precisely recognize the tokens within the Chinese names without prior knowledge.

The impact of historical context was also investigated, and it can be observed that for the decoder-only model, the addition of one historical context enhanced the percentage of the correct span by 1.3\% and the percentage of the correct entity by 2.0\%. This indicates that a longer context not only improved the model's ability to locate the entity but also promoted ASR accuracy as it ensured that the predicted entities were more coherent across utterances. For the encoder-decoder model, adding one historical context only increased the percentage of the correct span by 1.1\% and the percentage of the correct entity by 0.8\%. 

\begin{table}[h]
\caption{The classification of the entity predictions into different categories: Correct Span (Cor. Span), Correct Entity (Cor. Ent.), Error Span (Err. Span), Replacement (Rep.), and Omission (Omi.). The numbers are \textbf{percentages (\%)}.}
\label{tab:ner2}
\centering
\begin{tabular}{c | c | c c | c c c} 
\toprule
& & Cor. & Cor. & Err. & & \\
Method & His. & Span & Ent. & Span & Rep. & Omi.\\
\midrule
Conformer & 0 & 82.3 & 71.0 & 17.7 & 1.5 & 11.5 \\
\midrule
Decoder-& 0 & 88.7 & 77.1 & 11.3 & 1.3 & 5.6 \\
only & 1 & 90.0 & 79.1 & 10.0 & 1.1 & 5.2 \\
& 2 & \textbf{90.7} & \textbf{79.9} & \textbf{9.3} & \textbf{1.0} & \textbf{4.5} \\
\midrule
Encoder- & 0 & 88.4 & 77.4 & 11.6 & 1.3 & 5.8 \\
decoder & 1 & \textbf{89.5} & \textbf{78.2} & \textbf{10.5} & \textbf{1.2} & \textbf{5.2} \\
& 2 & 88.9 & 78.0 & 11.1 & 1.4 & 5.5 \\
\bottomrule
\end{tabular}
\end{table}

\section{Conclusion}

In this paper, we explored combining a speech encoder with an LLM for Chinese ASR and NER tasks using two different architectures: decoder-only and encoder-decoder. Under utterance-level evaluation, both architectures achieved significant improvements over the baseline Conformer model. We further compared these two approaches in terms of their capability to utilize long-form historical information. The long-form evaluations indicate that the decoder-only model benefited more from incorporating historical context than the encoder-decoder model. To explain this phenomenon, we conducted a comprehensive analysis of the gate values and attention scores, which revealed the superior ability of the decoder-only model to adjust the importance of speech and text modalities dynamically. For future works, we intend to conduct larger-scale experiments and evaluate our systems on more tasks. We hypothesize that the encoder-decoder model may have an advantage over the decoder-only approach in tasks where audio features are crucial, such as audio event detection. Therefore, we plan to investigate the potential of combining the two approaches to achieve superior performance on a wide range of tasks that require both high-level semantic and fine-grained acoustic information.

\bibliographystyle{IEEEtran}
\bibliography{mybib}

\end{document}